\documentclass[letterpaper]{article} 
\pdfoutput=1
\usepackage{aaai24}  
\usepackage{times}  
\usepackage{helvet}  
\usepackage{courier}  
\usepackage[hyphens]{url}  
\usepackage{graphicx} 
\usepackage{natbib}  
\usepackage{caption} 
\usepackage{algorithm}
\usepackage{algorithmic}
\usepackage{graphicx}
\usepackage{amsmath}
\usepackage{amssymb}
\usepackage{booktabs}
\usepackage{multirow}
\usepackage{makecell}
\usepackage{newfloat}
\usepackage{listings}
\DeclareCaptionStyle{ruled}{labelfont=normalfont,labelsep=colon,strut=off} 
\floatstyle{ruled}
\newfloat{listing}{tb}{lst}{}
\floatname{listing}{Listing}
\nocopyright

\title{TiC: Exploring Vision Transformer in Convolution}
\author{
    Song Zhang \equalcontrib \textsuperscript{\rm 1} 
    Qingzhong Wang \equalcontrib \textsuperscript{\rm 1},
    Jiang Bian \textsuperscript{\rm 1}, 
    Haoyi Xiong\thanks{Corresponding author} \textsuperscript{\rm 1}
}
\affiliations{
    \textsuperscript{\rm 1}Baidu Inc.\\
}

\usepackage{bibentry}

\begin{document}

\maketitle

\begin{abstract}

While models derived from Vision Transformers (ViTs) have been phonemically surging, pre-trained models cannot seamlessly adapt to arbitrary resolution images without altering the architecture and configuration, such as sampling the positional encoding, limiting their flexibility for various vision tasks.
%
%
For instance, the Segment Anything Model~(SAM) based on ViT-Huge requires all input images to be resized to 1024×1024. 
%
%
%
To overcome this limitation, we propose the Multi-Head Self-Attention Convolution (MSA-Conv) that incorporates Self-Attention within generalized convolutions, including standard, dilated, and depthwise ones. Enabling transformers to handle images of varying sizes without retraining or rescaling, the use of MSA-Conv further reduces computational costs compared to global attention in ViT, which grows costly as image size increases. 
Later, we present the Vision \underline{T}ransformer \underline{i}n \underline{C}onvolution (TiC) as a \emph{proof of concept} for image classification with MSA-Conv, where two capacity enhancing strategies, namely Multi-Directional Cyclic Shifted Mechanism and Inter-Pooling Mechanism,
have been proposed, through establishing long-distance connections between tokens and enlarging the effective receptive field.
%
%
%
%
%
%
%
%
%
%
Extensive experiments have been carried out to validate the overall effectiveness of TiC. 
%
%
Additionally, ablation studies confirm the performance improvement made by MSA-Conv and the two capacity enhancing strategies separately. Note that our proposal aims at studying an alternative to the global attention used in ViT, while MSA-Conv meets our goal by making TiC comparable to state-of-the-art on ImageNet-1K.
Code will be released at https://github.com/zs670980918/MSA-Conv.

\end{abstract}

\section{Introduction}

Images have become indispensable to our daily life, leading to massive image data generation and collection to enable ubiquitous applications in various domain~\cite{b1}. Vision transformer (ViT) architectures~\cite{b4} have emerged as one of the most promising ways to handle vast image data \cite{b14} for versatile pattern recognition \cite{b45} and image understanding tasks \cite{b46}. For instance, Segment Anything Model (SAM)~\cite{b2} with more than 100M learnable parameters, pre-trained on large-scale general-purpose instance segmentation and scene parsing datasets such as COCO \cite{b43} and ADE20K 
\cite{b44}, could perform zero-shot segmentation on images from various data sources without further tuning.



Specifically, in the ViT design, images are segmented into discrete non-overlapping patches, typically sized at 16 × 16 \cite{b5}. These patches are then treated as tokens, analogous to the tokens in NLP tasks~\cite{b3}, and integrated with positional encoding to represent coarse-grained spatial information. Subsequently, the data undergoes repeated standard transformer layers to model global relations \cite{b4}, ultimately enabling effective image classification.


While ViT-based models have achieved remarkable success in various vision tasks, they might not be able to be applied to high-resolution images without changing the architecture or configuration like re-sampling the positional encoding \cite{b41}, limiting the accessibility of models to various vision tasks. For instance, SAM based on ViT-Huge requires all input images to be resized to 1024×1024. In contrast, convolutional architectures like ResNet \cite{b6} are able to handle images of arbitrary resolutions. Furthermore, as the scale of input images increases, ViT-based models with global attention incur substantial computational overhead, hindering their ability to handle high-resolution images \cite{b8}.

To overcome these limitations and unlock the full potential of ViT for high-resolution images, we propose a novel approach that enhances the adaptability and computational efficiency of ViT. Our proposed Multi-Head Self-Attention Convolution (MSA-Conv) mechanism allows ViT to accept images with arbitrary resolution, eliminating the requirement of resizing or re-sampling the positional encoding, and facilitating seamless integration into various vision tasks. MSA-Conv leverages the power of convolutions, such as standard convolution \cite{b9}, dilated convolution \cite{b10}, and depthwise convolution \cite{b11}, which can be applied to arbitrary image resolution. 

\begin{figure}[!t]
\centering
\includegraphics[width=0.48\textwidth]{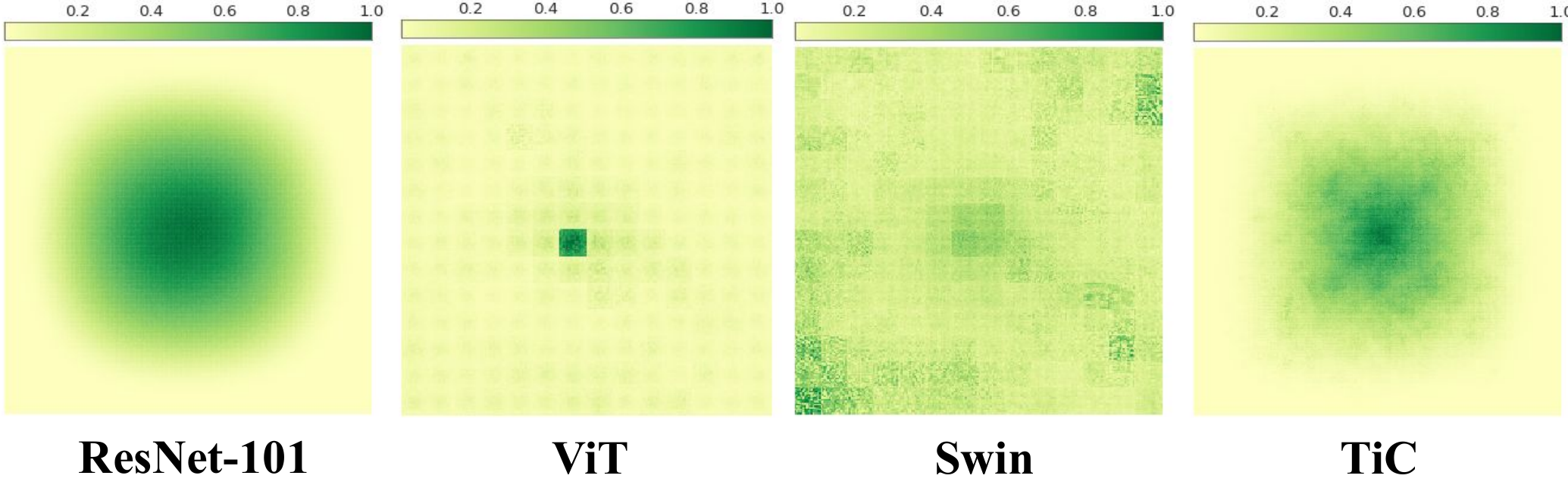}
\caption{Effective receptive field \cite{b42} of the mainstream backbones and TiC.}
\label{fig:Figure 0}
\vspace{-5mm}
\end{figure}

Further, during our exploration of constructing novel architectures with MSA-Conv, we observe that replacing global attention used in ViT with MSA-Conv would result in sub-optimal performance on the ImageNet-1K \cite{b13} image classification task. Such degradation, however, can be mitigated with the adoption of larger kernel sizes in MSA-Conv. This observation inspires us to study the effective receptive field problem, as MSA-Conv simply combining convolution with self-attention \cite{b12} would cause similar kernel size issues in convolutional neural networks~\cite{b42}. 
Hereby, in this paper, we present Vision Transformer in Convolution (TiC)--a proof-of-concept ViT design incorporating our proposed MSA-Conv with two advanced strategies as follows.
\begin{itemize}
    \item \emph{Multi-Directional Cyclic Shifted Mechanism:} Inspired by Swin \cite{b8}, TiC adopts a token shift mechanism to establish long-distance connections, employing four distinct manners to establish connections between diagonal position tokens for long-range connectivity. 
    
    \item \emph{Inter-Pooling Mechanism:} To avoid the additional computational overhead caused by the use of larger kernel sizes, TiC adopts an inter-pooling mechanism to reduce the size of the token map while enlarging the effective receptive field by pooling neighboring tokens. 
\end{itemize}
Above two strategies, together with patch downsample layers in TiC, could efficiently enlarge the effective receptive field and improve performance substantially, without significantly consuming computational costs. 
%
%
To understand the effectiveness of MSA-Conv and TiC over existing architectures, we visualize and compare the effective receptive fields obtained by four architectures, including ResNet-101, ViT, Swin and TiC.
%
Figure \ref{fig:Figure 0} illustrates the effective receptive fields of the four architectures on the same image. The effective receptive field of our newly proposed TiC demonstrates a resemblance to ResNet-101 \cite{b6}, which gives precedence to pixels near the center by assigning them larger weights. In contrast, ViT leans towards assigning maximum weight to the exact center, while the Swin shows a tendency to attribute greater importance to pixels distanced from the center. The observation back-ups the motivation of our proposal that incorporates convolution within transformers.



Comprehensive experimental evaluations have been conducted to ascertain the overarching efficacy of MVS-Conv. Furthermore, through ablation analyses, the enhancements introduced by MSA-Conv and the two capacity enhancing strategies have been individually validated. It is important to emphasize that our endeavor revolves around investigating an alternative to the global attention mechanism employed in ViT. MSA-Conv aligns with our objectives by rendering TiC competitive with state-of-the-art benchmarks on the ImageNet-1K dataset. Note that, though MSA-Conv and TiC are part of contributions claimed in this work, they might not be the best solutions in many of our experiments. However, both ablation studies and overall evaluation have proved the concepts of incorporating convolution within transformers with MSA-Conv, enlarging effective receptive fields via two strategies, and allowing images of arbitrary resolutions in TiC. Additional efforts are requested along this line of research for providing product-grade solutions.


\section{Related Work}
Convolutional-Based models have been at the forefront of computer vision for several decades \cite{b7, b15, b16}. The seminal work of AlexNet \cite{b7} catapulted CNNs into the mainstream, leading to the development of deeper and more effective Conv-Based architectures, including VGG \cite{b15}, GoogleNet \cite{b16}, ResNet \cite{b6}, DenseNet \cite{b17}, HRNet \cite{b18}, and EfficientNet \cite{b19}. Furthermore, research efforts have been devoted to enhancing individual convolution layers, resulting in innovations such as depthwise convolution \cite{b11} and deformable convolution \cite{b30}.

In recent years, ViT models have garnered increasing attention and demonstrated significant success in computer vision tasks \cite{b20}. Pioneering work like ViT \cite{b4} directly applied the transformer architecture from NLP to image classification using image patches as input. Subsequent studies, such as Swin \cite{b8} and CSWin \cite{b21}, have further explored self-attention within shifted local windows, and PVT \cite{b22} introduced a pyramid structure to generate multi-scale feature maps for pixel-level dense prediction tasks. Moreover, CPVT \cite{b23} and CvT \cite{b3} leveraged a convolutional projection into transformers.

However, current pre-trained models \cite{b24, b25} cannot directly apply to high-resolution images without changing the architecture and configuration like re-sampling the positional encoding \cite{b41}, limiting the flexibility compared to Conv-Based architectures. To capitalize on the strengths of both approaches, we draw inspiration from the sliding window mechanism of convolution and integrate it into the ViT-based models. This fusion aims to enhance performance and flexibility in computer vision applications, leveraging the advantages of Conv-Based architectures and ViTs simultaneously.


\begin{figure*}[!t]
\centering
\includegraphics[width=1\textwidth]{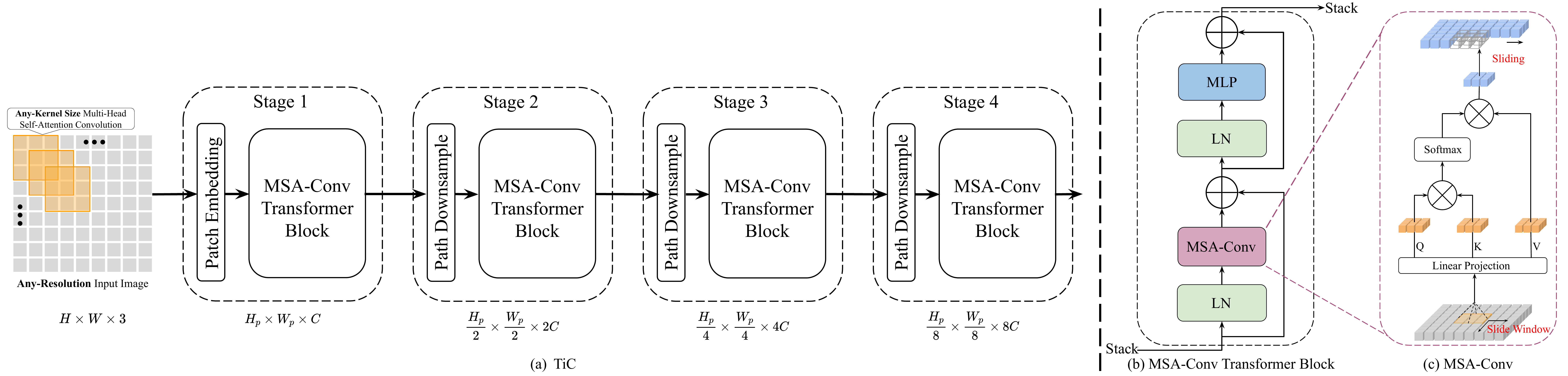}
\caption{(a) The architecture of TiC; (b) the architecture of MAS-Conv Transformer Block; (c) the illustration of MSA-Conv. }
\label{fig:Figure 1}
\vspace{-5mm}
\end{figure*}

\section{Methodology}
While the major claim of this work is the proposal of Multi-head Self-Attention Convolution~(MSA-Conv), as an alternative to the global attention used in ViT. We also introduce Vision \underline{T}ransformer \underline{i}n \underline{C}onvolution (TiC) as a \emph{proof of concept} solution leveraging MSA-Conv for image classification of arbitrary resolutions. Here, we first present the overall design of TiC, then specify the design of MSA-Conv in details.

\subsection{Framework}

\paragraph{Overall Design of TiC}
As shown in Figure~\ref{fig:Figure 1}, TiC adopts a hierarchical architecture consisting of several MSA-Conv Transformer Blocks and Patch Downsample layers, where every block leverages an any-kernel size multi-head self-attention convolution facilitating feature extraction and learning from images of diverse resolutions in a transformer layer.
%
%
Specifically, in ``Stage 1'' of TiC, the initial MSA-Conv Transformer Block processes the token features from the patch embedding module, preserving the full spatial resolution of $H \times W$ tokens. To introduce a hierarchical representation, the patch downsample layer progressively reduces the number of tokens as the network depth increases. Specifically, the first downsample layer concatenates the features of each $2 \times 2$ neighboring patch group, resulting in 4C-dimensional features. These features are then projected by a linear layer to 2C dimensions, effectively reducing the number of tokens by a factor of 4 ($2\times$ downsampling of spatial resolution). Subsequent MSA-Conv Transformer Blocks operate on this lower resolution feature map of $H_{p} \times W_{p}$~($H_{p} = \frac{H}{4}, W_{p} = \frac{W}{4}$), constituting ``Stage 2'' of TiC. To the end, such hierarchical design effectively combines multi-scale feature learning with long-range modeling.

\paragraph{MSA-Conv Transformer Block}
More specifically, inside every MSA-Conv Transformer Block, the standard multi-head self-attention~(MSA) module used by the vanilla Transformer block is replaced by an any-kernel size multi-head self-attention convolution~(MSA-Conv), while the remaining layers remain unchanged. As depicted in Figure \ref{fig:Figure 1}, an MSA-Conv Transformer Block consists of the MSA-Conv module, followed by a 2-layer MLP \cite{b26} with GELU \cite{b27} nonlinearity in between. LayerNorm (LN) \cite{b28} layers are applied before each MSA-Conv module and each MLP, with a residual connection applied after each module.

\paragraph{Feature Extraction and Outputs}
The process involving MSA-Conv Transformer Blocks and Patch Downsample layers is repeated twice more, constituting ``Stage 3'' and ``Stage 4'', which yield feature maps at spatial resolutions of $\frac{H}{16} \times \frac{W}{16}$ and $\frac{H}{32} \times \frac{W}{32}$, respectively. This hierarchical architecture allows the network to generate a multi-scale feature representation, comparable to resolutions typically produced by convolutional backbones such as VGG \cite{b15} and ResNet \cite{b6}. As a result, the proposed architecture can easily replace backbone networks in existing methods for various computer vision tasks.

\begin{figure}[!t]
\centering
\includegraphics[width=0.5\textwidth]{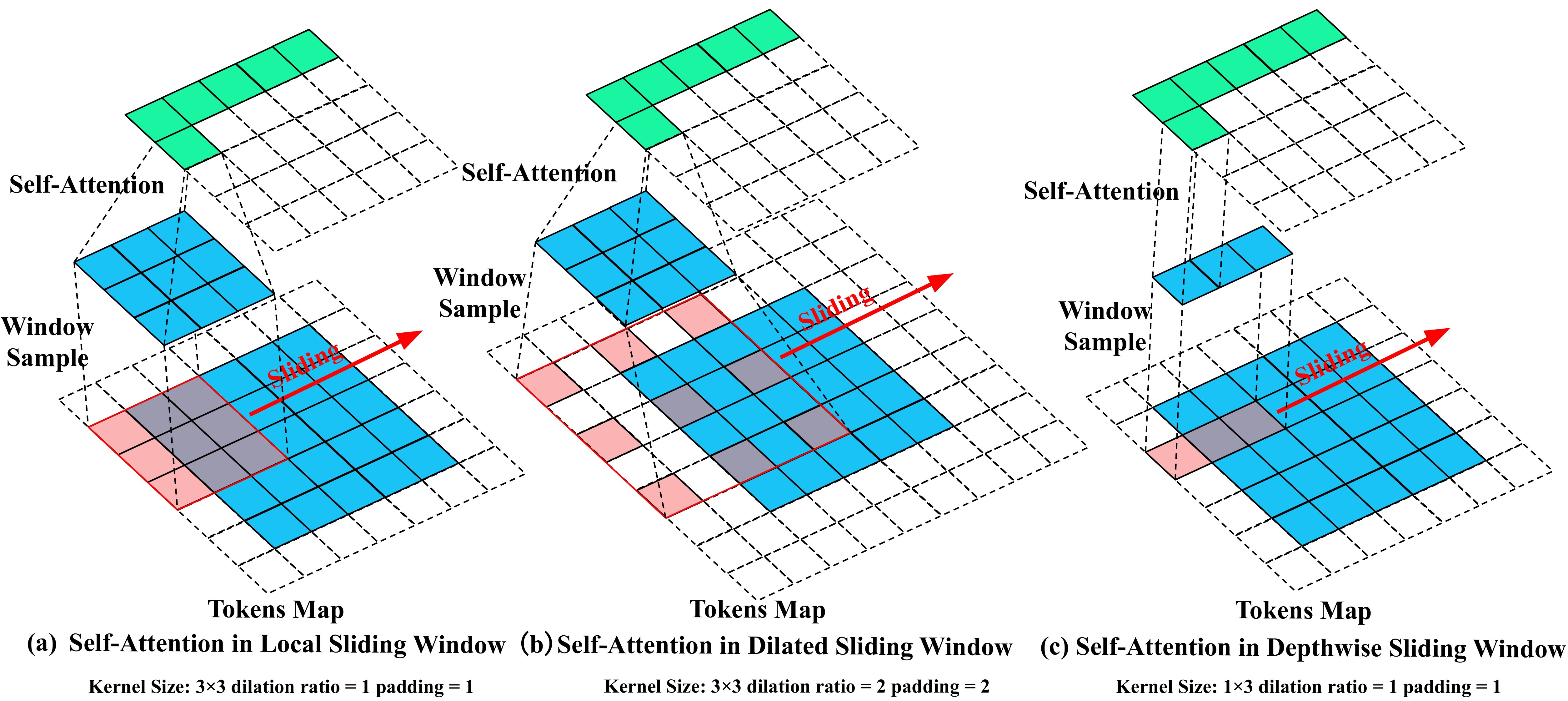}
\caption{Illustration of MSA-Conv operations. (a): the self-attention w/. local sliding window; (b): w/. dilated sliding window; (c): w/. depthwise sliding window.}
\label{fig:Figure 2}
\vspace{-5mm}
\end{figure}


\subsection{Multi-Head Self-Attention Convolution}
\label{sec:msa_conv}
The Vision Transformer architecture \cite{b4} and its adaptation for image classification employ global self-attention, where the relationships between a token and all other tokens are computed. However, the global computation leads to quadratic complexity concerning the number of tokens, rendering it unsuitable for many vision tasks that require an extensive set of tokens for dense prediction or to represent high-resolution images. To address this limitation, the Swin-Transformer \cite{b8} proposes local self-attention, wherein self-attention is computed within local windows, reducing computational demand. While these methods demonstrate excellent performance in image classification, they process different resolutions images need to re-sample the positional encoding~(ViT \cite{b4}) or adjust the configuration of architecture~(Swin-Transformer \cite{b8}). 

To address these challenges and enable the flexible processing of any-resolution input images while fully utilizing local attention within an any-kernel size window, we introduce the Multi-Head Self-Attention Convolution (MSA-Conv). This mechanism allows for self-attention computation within a local sliding window of any kernel size, akin to the convolutional mechanism. We have implemented MSA-Conv as a convolutional layer, which further enhances its ease of integration within various vision architectures.

\paragraph{Self-Attention in Local Sliding Window} Local attention has demonstrated its effectiveness in the Swin-Transformer \cite{b8} architecture. In our proposed attention pattern, we employ a combination of fixed-size window attention surrounding each token and flexible-size window attention across different layers. The use of multiple stacked layers of such window attention enables a large receptive field, granting top layers access to all input locations and the capacity to build representations incorporating information across the entire input, akin to CNNs \cite{b29}. The local sliding window illustration is provided in Fig. \ref{fig:Figure 2}(a). Let us assume that each window contains $M \times M$ patches in a layer, and the kernel size of MSA-Conv is $K \times K$. The computational complexity of a local sliding window-based MSA on an image with $h \times w$ patches is depicted as follows in Eq. \eqref{eq:equation_1} $\Omega(\mathrm{MSA-Conv})$:
\begin{equation}
    \label{eq:equation_1}
    \begin{aligned}
    & \text{TiC: } \Omega(\mathrm{MSA-Conv})=4 h w C^2+K^{2} h w  C \\ 
    & \text{ViT: } \Omega(\mathrm{MSA})=4 h w C^2+2(h w)^2 C, \\
    & \text{Swin-T: } \Omega(\mathrm{W}-\mathrm{MSA})=4 h w C^2+2 M^2 h w C \\ 
    \end{aligned}
\end{equation}
Furthermore, we have conducted a comparative analysis of computational complexity with ViT \cite{b4} and Swin-Transformer~(Swin-T) \cite{b8} in Eq. \eqref{eq:equation_1}. As observed, the global self-attention computation~($\Omega(\mathrm{MSA})$) generally becomes impractical for large $hw$ dimensions, whereas the window-based self-attention~($\Omega(\mathrm{MSA-Conv}), \Omega(\mathrm{W}-\mathrm{MSA})$) exhibits scalability. Both TiC and Swin-T demonstrate similar computational complexity, benefiting from the sliding window mechanism. Additionally, our MSA-Conv introduces the advantage of supporting any-resolution input images compared to other methods while maintaining similar efficiency.

\paragraph{Self-Attention in Dilated Sliding Window} We aim for our MSA-Conv to function as a comprehensive convolutional layer, encompassing both the normal sliding window capability and a widely used dilated capability to enhance the receptive field of the local window. To achieve a broader receptive field without increasing computational overhead, we introduce the concept of "dilation" to the sliding window. This concept draws parallels with dilated CNNs \cite{b10}, where gaps of size dilation $d$ are introduced within the window. Assuming a fixed $d$ and $w$ for all layers, the resulting receptive field becomes $d \times h \times w$, potentially reaching hundreds of patch tokens even for small dilation values. In the context of multi-head self-attention, each self-attention head computes distinct attention scores. Inspired by this, we implement different dilation configurations per head to improve performance. This allows some heads without dilation to focus on local context, while others with dilation focus on the larger context. The illustration of the dilated sliding window is depicted in Fig. \ref{fig:Figure 2}(b).

\paragraph{Self-Attention in Depthwise Sliding Window} Figure \ref{fig:Figure 2}(c) illustrates the mechanism of our depthwise sliding window. Based on this mechanism, our MSA-Conv can accommodate any-kernel size window~(e.g., $1 \times K$, $K \times 1$, $K \times K$) for self-attention computation. Consequently, to increase the effective receptive field without significantly escalating computation, larger kernel sizes can be adopted to enhance performance. This concept shares similarities with depthwise CNNs \cite{b11}, where each input channel is convolved with a distinct kernel. Assuming the MSA-Conv possesses four self-attention heads, different kernel sizes can be employed for each self-attention head. In line with the above different dilation configurations per head, we also select one head to utilize a larger kernel size, resulting in a larger receptive field for focusing on the broader context.

\begin{figure}[!t]
\centering
\includegraphics[width=0.5\textwidth]{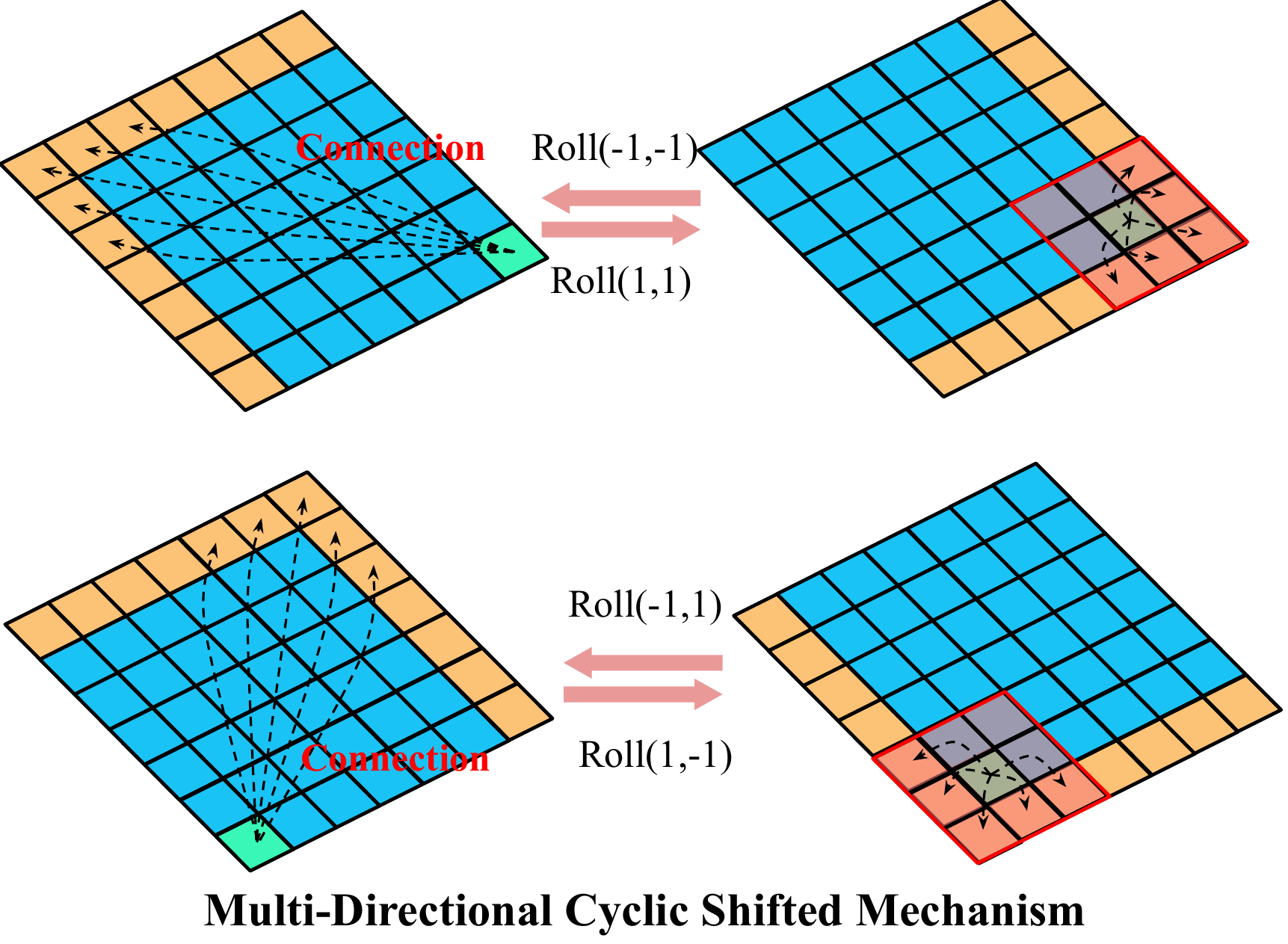}
\caption{Illustration of overall Multi-Directional Cyclic Shifted Mechanism. Roll refers to \cite{b32}.}
\label{fig:Figure 3}
\vspace{-5mm}
\end{figure}

\subsection{Effective Receptive Field Enlargement Strategies}
The local sliding window-based MSA-Conv module lacks connections among patch tokens in remote locations, limiting its modeling power. While the dilated sliding window and depthwise sliding window have enhanced the receptive field to some extent, they are not potent enough to establish connections between diagonal position tokens, essential for obtaining global task-specific representations. To address this limitation, we have devised strategies to establish long-distance connections, thereby enhancing the capability of task-specific information perception in our model.

\paragraph{Multi-Directional Cyclic Shifted Mechanism} 
The local sliding window-based self-attention module lacks the capability of establishing long-distance connections across patch tokens, leading to limitations in its modeling power. Taking inspiration from Swin-Transformer \cite{b8}, we propose a multi-directional cyclic shifted mechanism to introduce cross-token connections without incurring additional computation. This mechanism randomly cyclically shifts in four directions, enabling patch tokens that are difficult to relate due to the window size to be included in the same window. As a result, the receptive field is increased, thereby enhancing the capability of task-specific information perception in our model. The illustration of this mechanism is provided in Fig. \ref{fig:Figure 3}. In our experiments, we ensure full interaction of information by setting the size of cyclic shifted to half of the window size, guaranteeing that a portion of the patch tokens originates from long-distance connections.

\paragraph{Inter-Pooling Mechanism} To increase the receptive field for capturing more contextual information while simultaneously reducing computational complexity, we introduce an inter-pooling mechanism. This mechanism employs convolutions to pool the patch tokens, enabling the aggregation of additional neighboring context information within a single window. The formulation of our inter-pooling mechanism is defined as follows in Eq. \ref{eq:equation_2}:

\begin{equation}
    \label{eq:equation_2}
    \begin{aligned}
    & \text{Q: } Rerrange(\mathbf{Q}, f) \\ 
    & \text{K: } Repeat(\mathbf{K} \odot w_{1}, \mathbf{Q}.shape) \\ 
    & \text{V: } Repeat(\mathbf{V} \odot w_{2}, \mathbf{Q}.shape) \\ 
    \end{aligned}
\end{equation}

\noindent where $f$ represents pooling factor, $\odot$ represents the Hadamard multiplication, $ w_{i} $ represents the weights of the convolutional layers. Specifically, for the query~($\mathbf{Q}$), we employ the Rerrange operation~(from the einops library \cite{b31}) to reshape $\mathbf{Q}$. The shape transformation formulation is as follows: $[B, C, H, W] \rightarrow [B\times f^{2}, C, \frac{H}{f}, \frac{H}{f}]$. As for the key~($\mathbf{K}$) and value~($\mathbf{V}$), we utilize separate convolutions to downsample~(aggregate information) each item~(shape: $[B, C, H, W] \rightarrow [B, C, \frac{H}{f}, \frac{H}{f}]$), and then adopt the Repeat operation~(from the einops library \cite{b31}) to reshape the shape. The shape transformation formulation is: $[B, C, \frac{H}{f}, \frac{H}{f}] \rightarrow [B\times f^{2}, C, \frac{H}{f}, \frac{H}{f}]$. Through inter-pooling, we can efficiently aggregate information while simultaneously increasing the batch size for CUDA, achieving computational acceleration.

\section{Experiments}
In this section, we discuss the enhancement of model performance by effective receptive fields and perform related ablation experiments. And we also present an efficiency analysis of the proposed TiC architecture against previous state-of-the-art methods to validate the flexibility of our proposed MSA-Conv. Additionally, we conduct extensive experiments on ImageNet-1K image classification \cite{b13} to validate the effectiveness of our MSA-Conv based on TiC.

\begin{figure}[!t]
\centering
\includegraphics[width=0.5\textwidth]{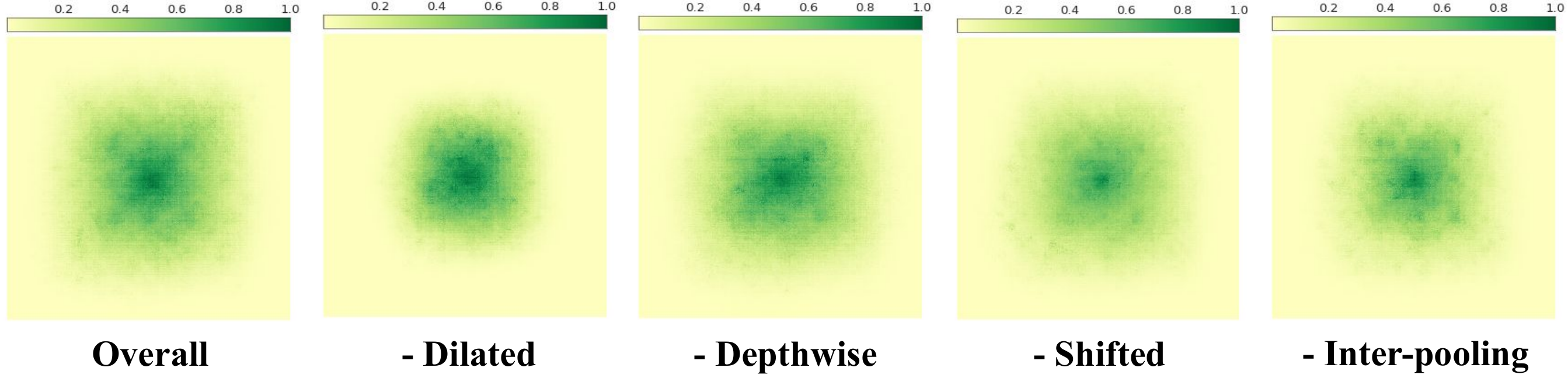}
\caption{Illustration of the Impact of Different Modules on the Effective Receptive Field.}
\label{fig:Figure 4}
\vspace{-5mm}
\end{figure}

\subsection{Effectiveness on Receptive Field Enlargement}
In this section, we discuss the impact of the improvement of the receptive field brought about by different strategies on performance. Specifically, we conducted experiments based on ImageNet-1K image classification \cite{b13}. The ablation study of each module is reported in Table \ref{tab:Table.3}. And the effective receptive field of each module is shown in Figure. \ref{fig:Figure 4}.

\begin{table}[!t]
    \centering
    \footnotesize
    \begin{tabular}{c|ccc}
    \toprule
        & \makecell{Backbone} & \makecell{ImageNet \\ Top-1 acc~(\%)} & \makecell{ImageNet \\ Top-5 acc~(\%)} \\ \hline
        - dilated & TiC-B & 76.47 & 92.37 \\
        overall & TiC-B & 79.00 & 94.31 \\ \hline
        - depthwise & TiC-B & 78.72 & 94.22 \\
        overall & TiC-B & 79.00 & 94.31 \\ \hline
        - cyclic shifted & TiC-B & 78.46 & 93.78 \\
        overall & TiC-B & 79.00 & 94.31 \\ \hline
        - inter-pooling & TiC-B & 77.32 & 93.30 \\
        overall & TiC-B & 79.00 & 94.31 \\ 
    \bottomrule
    \end{tabular}
    \caption{Ablation study on ImageNet-1K. '-' represents removing the module from the overall TiC.}
    \label{tab:Table.3}
    \vspace{-5mm}
\end{table}

\paragraph{Dilated Sliding Window} As illustrated in Figure \ref{fig:Figure 4}, it is evident that the removal of the dilated sliding window leads to a reduction in the effective receptive field of the model compared to the overall TiC architecture (Figure \ref{fig:Figure 4} TiC vs. Figure \ref{fig:Figure 4} -Dilated). Furthermore, the implications of dilated sliding window ablations on classification performance are detailed in Table \ref{tab:Table.3}. Notably, TiC incorporating the dilated sliding window surpasses its local sliding window counterpart by a margin of 2.53 in terms of top-1 accuracy on the ImageNet-1K. These results substantiate the efficacy of leveraging the dilated sliding window to amplify the effective receptive field and foster connections between patch tokens spanning longer distances.

\paragraph{Depthwise Sliding Window} As depicted in Figure \ref{fig:Figure 4}, the utilization of the depthwise sliding window results in a discernible augmentation of the effective receptive field. Although the expansion of the effective receptive field through the depthwise sliding window is comparatively smaller when contrasted with the effects of the dilated sliding window, we posit that this discrepancy could be attributed to the application of the depthwise sliding window solely on a single head within the Multi-Head configuration. This observation is further substantiated through quantitative assessment, as highlighted in Table \ref{tab:Table.3}. Notably, TiC incorporating the depthwise sliding window achieves an increment of 0.28 in top-1 accuracy on the ImageNet-1K dataset when compared to its local sliding window counterpart. These outcomes underscore the efficacy of leveraging the depthwise sliding window to amplify the perceptual field and facilitate the establishment of connections between patch tokens spanning longer distances.

\paragraph{Multi-Directional Cyclic Shifted Mechanism} As
illustrated in Figure \ref{fig:Figure 4}, our qualitative results reveal that the incorporation of the shifted mechanism generates a discernible elevation in attention towards edge tokens. This shift is perceptible through the darker coloration and extended range of the effective receptive field's edge, consequently fostering the establishment of connections between tokens situated at distant positions. This augmentation is anticipated to yield an enhancement in model performance. This empirical observation is corroborated by the quantitative outcomes, as detailed in Table \ref{tab:Table.3}. Specifically, our findings substantiate the efficacy of amplifying attention towards edge tokens in bolstering model performance. Notably, TiC integrating the multi-directional cyclic shifted mechanism exhibits a notable superiority over its counterpart devoid of this mechanism, showcasing a commendable increment of 0.54 in top-1 accuracy on the ImageNet-1K dataset.

\paragraph{Inter-Pooling Mechanism} As depicted in Figure \ref{fig:Figure 4}, a discernible enhancement in the effective receptive field is evident the implementation of inter-pooling. This augmentation is attributed to the aggregation of neighboring tokens within the same local window for self-attention. The consequence is an increase in the participation of tokens within self-attention, consequently fostering an expansion of the effective receptive field. The efficacy of our inter-pooling strategy is further substantiated by quantitative results. Quantitatively, the results demonstrate the superiority of TiC integrated with the inter-pooling mechanism over its counterpart lacking this mechanism. Specifically, TiC endowed with the inter-pooling mechanism showcases a noteworthy advancement of 1.68 in top-1 accuracy on the ImageNet-1K dataset. This numerical validation underscores the potency of the inter-pooling mechanism in enlarging the receptive field and thereby enhancing model performance.


In summary, drawn from the comprehensive ablation study analysis, it is evident that the effective receptive field plays a pivotal role in model performance. Therefore, enhancing the effective receptive field stands as a pivotal avenue for improving model performance in future endeavors.

\begin{figure}[!t]
\centering
\includegraphics[width=0.52\textwidth]{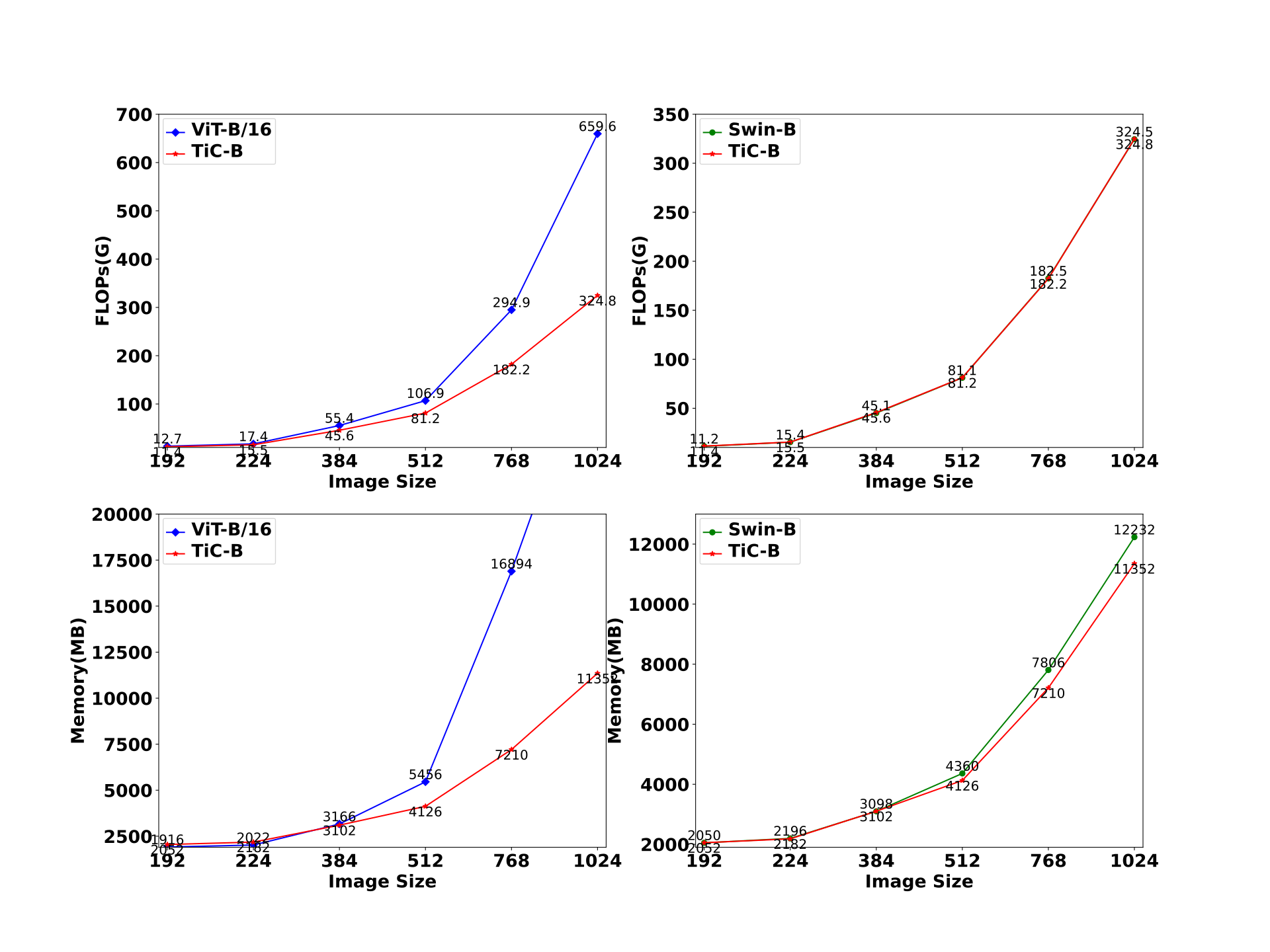}\vspace{-3mm}
\caption{Efficiency Analysis. Comparative evaluation of FLOPs and memory usage between TiC and mainstream ViT-based models across various image resolutions in the inference phase.}\vspace{-3mm}
\label{fig:Figure 5.1}
\end{figure}

\subsection{Efficiency under Varying Image Resolutions}
In this section, in order to further verify the flexibility and effectiveness of MSA-Conv on arbitrary resolution images, we compared the efficiency of various ViT-based models corresponding to different resolutions on ImageNet-1k in the inference phase. Moreover, we also discuss the runtime and memory of different implementations of TiC’s MSA-Conv in the inference phase. 

Significantly, during our experimentation phase, a notable observation arose concerning the adaptability of ViT-B/16 and Swin-B to varying image resolutions. In the specific context of ViT-B/16 trained on images with dimensions of $224 \times 224$, conducting inference on images of dissimilar resolutions necessitates the resampling of positional encoding to ensure proper utilization. Similarly, Swin-B encounters limitations related to the interplay between window size and image size. For instance, for an image of size $224 \times 224$, Swin-B mandates a window size adjustment to 7, while for an image of dimensions $192 \times 192$, the optimal window size becomes 6. This inflexibility in adaptability is a discernible drawback in these architectures. In contrast, our MSA-Conv architecture leverages the advantages of sliding window mechanisms and potential positional encoding, endowing it with the capability to seamlessly function across diverse image resolutions without necessitating any modifications.

\begin{figure}[!t]
\centering
\includegraphics[width=0.52\textwidth]{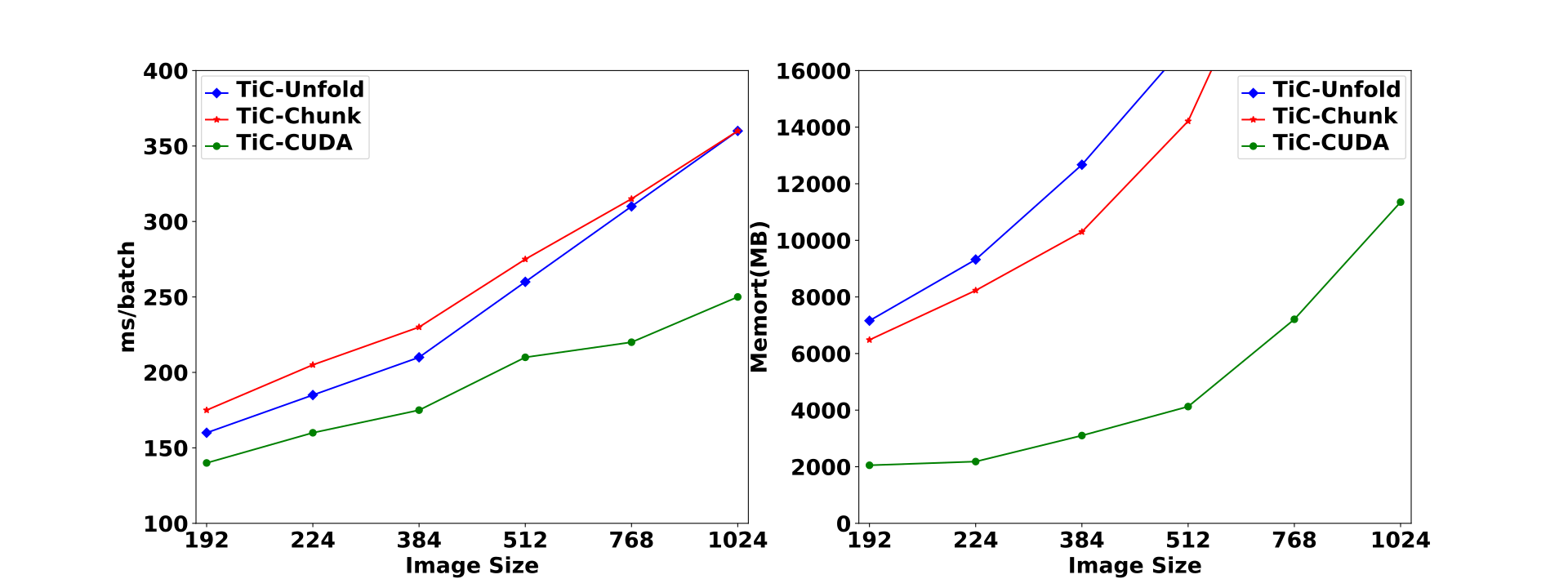}\vspace{-3mm}
\caption{Illustration of efficiency analysis of different implementations of MSA-Conv in the inference phase.}\vspace{-3mm}
\label{fig:Figure 6.1}
\end{figure}

As shown in Figure \ref{fig:Figure 5.1}, we can observe that our approach exhibits a distinct advantage over ViT with global attention in terms of FLOPs as image resolutions escalate. This observation is in alignment with the theoretical computational formulation described in Eq. \eqref{eq:equation_1}. Notably, as the image dimensions reach $1024 \times 1024$, the FLOPs of ViT-B/16 surge to almost twice that of TiC. This discrepancy underscores TiC's computational efficiency in contrast to ViT-B/16. Furthermore, the escalating memory usage of ViT-B/16 due to its global attention computational mechanism is of paramount significance. The involvement of an increasing number of patch tokens in computation, which is exponentiated by the expanding image size, amplifies the memory usage in ViT-B/16. Conversely, our method capitalizes on the local sliding window mechanism for self-attention computation. Consequently, the memory usage of TiC grows linearly in relation to the image size, offering a more favorable trade-off between memory utilization and image dimensions. Comparison with Swin-B, which is also based on local attention, showcases a parallel performance in terms of FLOPs, a correspondence that aligns with the predictions posited by Eq. \eqref{eq:equation_1}. Furthermore, leveraging the benefits of CUDA programming's resource scheduling capabilities, as the image scale keeps increasing TiC will gradually be lower than Swin-B in terms of memory consumption, e.g., $1024 \times 1024$, TiC: 11352MB vs. Swin-B 12232MB.

In addition, we conducted comparisons involving time and memory utilization across different implementations of TiC's MSA-Conv. Our TiC implementation encompasses three variants: 1. TiC-Unfold, 2. TiC-Chunk, and 3. TiC-CUDA. TiC-Unfold employs the nn.Unfold interface \cite{b32} from PyTorch to mimic a sliding window for value extraction, followed by self-attention computations using pure Python code. TiC-Chunk, an enhancement of TiC-Unfold, processes values from nn.Unfold in batches to optimize memory consumption through pure Python code. TiC-CUDA relies on CUDA programming \cite{b33}, utilizing C++ to implement underlying convolution calculation for forward and backward propagation, and leveraging CUDA for computational acceleration. In Figure \ref{fig:Figure 6.1}, it is evident that our TiC-CUDA leverages CUDA's computational advantages, surpassing the processing speed of both TiC-Unfold and TiC-Chunk. Furthermore, TiC-CUDA remarkably reduces memory consumption compared to its counterparts through matrix operations optimized via CUDA programming. CUDA's inherent memory management capabilities further enhance the efficiency of our TiC-CUDA for processing tasks.


In summary, grounded in the comprehensive experimental analysis presented above, our assertion is that MSA-Conv exhibits heightened flexibility and efficiency in accommodating images of any-resolutions when juxtaposed with prevailing mainstream ViT-based models.

\subsection{Effectiveness on Downstream Tasks}
To validate the effectiveness of our MSA-Conv, we evaluate the TiC framework on the ImageNet-1K image classification \cite{b13}. Our performance metric is the top-1 accuracy derived from a single crop. In the experimental setting, we use similar configurations as Swin-Transformer \cite{b8}.


\begin{table}[!t]
    \centering
    \footnotesize
    \begin{tabular}{c|cc|cc}
    \toprule
        Methods & \makecell{Image \\ size} & FLOPs~(G)  & \makecell{ImageNet \\ Top-1 acc~(\%)} & \makecell{ImageNet \\ Top-5 acc~(\%)} \\ \hline
        ViT-B/16 & $224^{2}$ & 55.4G & 76.45 & 93.08 \\ 
        Swin-B & $224^{2}$ & 15.4G & 80.48 & 95.21 \\ 
        TiC-B & $224^{2}$ & 15.5G & 79.00 & 94.31 \\ 
    \bottomrule
    \end{tabular}
    \caption{Comparison of different backbones from scratch training on ImageNet-1K for classification. Throughput is measured using the GitHub repository of \cite{b39} and 8x V100-16G GPU, following \cite{b35}.}\vspace{-5mm}
    \label{tab:Table.1}
\end{table}

\paragraph{Results with ImageNet-1K from scratch training} Table \ref{tab:Table.1} showcases performance comparisons of our TiC-B architecture against prevalent Transformer-based backbones. These evaluations are carried out through ImageNet-1K training from scratch within our environment. In contrast to established state-of-the-art Transformer-based models, specifically ViT-B/16 and Swin-B, our TiC-B demonstrates a substantial enhancement in top-1 accuracy on ImageNet-1K, surpassing ViT-B by 2.55 and yielding a comparable outcome of 79 against Swin-B's 80.48.

\paragraph{Results with ImageNet-1K pre-training and fine-tuning} We further perform pretraining of our TiC on the ImageNet1K dataset. Fine-tuned outcomes for ImageNet-1K image classification are presented in Table \ref{tab:Table.2}. When juxtaposed with prevailing state-of-the-art Transformer-based architectures, such as ViT-B/16 (81.79) and Swin-B (83.31), our TiC-B has achieved a commendably competitive top-1 accuracy of 82.71 on the ImageNet-1K benchmark. Specifically, our TiC-B surpasses ViT-B by 0.92 and yields a comparable outcome of 82.71 against Swin-B 83.31. These findings suggest that there remains ample potential for further enhancements in our model's performance.


In summation, drawn from the outcomes of the above experiments, it is evident that MSA-Conv demonstrates performance on par with existing approaches in downstream image classification. Moreover, its substantial potential as versatile and overarching ViT-based models for future applications is notably discernible.

\begin{table}[!t]
    \centering
    \footnotesize
    \begin{tabular}{c|c|cc}
    \toprule
        Methods & \makecell{Image \\ size}  & \makecell{ImageNet \\ Top-1 acc~(\%)} & \makecell{ImageNet \\ Top-5 acc~(\%)} \\ \hline
        ViT-B/16 & $224^{2}$ & 81.79 & 95.75 \\ 
        Swin-B & $224^{2}$ & 83.31 & 96.36 \\ 
        TiC-B & $224^{2}$ & 82.71 & 96.07\\ 
    \bottomrule
    \end{tabular}
    \caption{Comparison of different backbones pretraining on ImageNet-1K and finetuning on ImageNet-1K. For the sake of fairness, we maintain uniformity in our experimental environment and configuration.}\vspace{-5mm}
    \label{tab:Table.2}
\end{table}


\section{Discussion}
Despite matching state-of-the-art-methods on ImageNet-1K tasks \cite{b13}, our model reveals areas for improvement as follows.
\begin{itemize}

\item\textbf{Efficiency}: the C++/CUDA implementation of MSA-Conv presents inefficiencies because of computational overheads from intermediate value generation during $\mathbf{Dot}$ propagation stages. This may be mitigated by single-process computation of $\mathbf{Q},\mathbf{K},\mathbf{V}$ gradients, based on the backward gradient $\mathbf{\tilde{O}}$.
\item \textbf{Extensibility}: Our previous section outlined MSA-Conv implementations in standard, dilated, and depthwise configurations. We suspect there would be an opportunity for MSA-Conv to include deformable convolution. This extension might allow self-attention to autonomously choose computational patches, leading to prospective enhancements in model performance. Additionally, considering implementation methods and effective receptive field boundaries, there seems to be potential for modular MSA-Conv within CNNs. 
\end{itemize}
Thus, we advocate for future research in above areas. 

\section{Conclusion}

This study presents a transformative proposal, MSA-Conv, that incorporates Self-Attention within generalized convolutions. This approach offers adaptability to Vision Transformers (ViTs) for handling images of variable resolutions without the need for retraining or rescaling, effectively addressing limitations faced in past ViT-based models like SAM. The introduction of Vision Transformer in Convolution (TiC) served as a proof of concept, demonstrating the pivotal role played by MSA-Conv, alongside two capacity enhancing strategies, in boosting the overall effectiveness of TiC. The efficacy of TiC was substantiated through several extensive experiments and ablation studies. This pursuit of finding a viable alternative to the global attention used in ViT found success in MSA-Conv, making TiC a comparable baseline against leading benchmarks on ImageNet-1K. Though TiC might not be the best solution in experimental comparisons, however it sheds light on the future research directions of incorporating transformers in convolution, enlarging effective receptive fields for enhanced capacity, and handling images of arbitrary/variable resolutions.

\bibliography{aaai24}

\begin{thebibliography}{58}
\providecommand{\natexlab}[1]{#1}

\bibitem[{Ba, Kiros, and Hinton(2016)}]{b28}
Ba, J.~L.; Kiros, J.~R.; and Hinton, G.~E. 2016.
\newblock Layer normalization.
\newblock \emph{arXiv preprint arXiv:1607.06450}.

\bibitem[{Bommasani et~al.(2021)Bommasani, Hudson, Adeli, Altman, Arora, von
  Arx, Bernstein, Bohg, Bosselut, Brunskill et~al.}]{b14}
Bommasani, R.; Hudson, D.~A.; Adeli, E.; Altman, R.; Arora, S.; von Arx, S.;
  Bernstein, M.~S.; Bohg, J.; Bosselut, A.; Brunskill, E.; et~al. 2021.
\newblock On the opportunities and risks of foundation models.
\newblock \emph{arXiv preprint arXiv:2108.07258}.

\bibitem[{Bouvrie(2006)}]{b9}
Bouvrie, J. 2006.
\newblock Notes on convolutional neural networks.

\bibitem[{Chollet(2017)}]{b11}
Chollet, F. 2017.
\newblock Xception: Deep learning with depthwise separable convolutions.
\newblock In \emph{Proceedings of the IEEE conference on computer vision and
  pattern recognition}, 1251--1258.

\bibitem[{Chu et~al.(2021)Chu, Tian, Zhang, Wang, Wei, Xia, and Shen}]{b23}
Chu, X.; Tian, Z.; Zhang, B.; Wang, X.; Wei, X.; Xia, H.; and Shen, C. 2021.
\newblock Conditional positional encodings for vision transformers.
\newblock \emph{arXiv preprint arXiv:2102.10882}.

\bibitem[{Clancey(1979)}]{c:79}
Clancey, W.~J. 1979.
\newblock \emph{{Transfer of Rule-Based Expertise through a Tutorial
  Dialogue}}.
\newblock {Ph.D.} diss., Dept.\ of Computer Science, Stanford Univ., Stanford,
  Calif.

\bibitem[{Clancey(1983)}]{c:83}
Clancey, W.~J. 1983.
\newblock {Communication, Simulation, and Intelligent Agents: Implications of
  Personal Intelligent Machines for Medical Education}.
\newblock In \emph{Proceedings of the Eighth International Joint Conference on
  Artificial Intelligence {(IJCAI-83)}}, 556--560. Menlo Park, Calif: {IJCAI
  Organization}.

\bibitem[{Clancey(1984)}]{c:84}
Clancey, W.~J. 1984.
\newblock {Classification Problem Solving}.
\newblock In \emph{Proceedings of the Fourth National Conference on Artificial
  Intelligence}, 45--54. Menlo Park, Calif.: AAAI Press.

\bibitem[{Clancey(2021)}]{c:21}
Clancey, W.~J. 2021.
\newblock {The Engineering of Qualitative Models}.
\newblock Forthcoming.

\bibitem[{Dai et~al.(2017)Dai, Qi, Xiong, Li, Zhang, Hu, and Wei}]{b30}
Dai, J.; Qi, H.; Xiong, Y.; Li, Y.; Zhang, G.; Hu, H.; and Wei, Y. 2017.
\newblock Deformable convolutional networks.
\newblock In \emph{Proceedings of the IEEE international conference on computer
  vision}, 764--773.

\bibitem[{Deng et~al.(2009)Deng, Dong, Socher, Li, Li, and Fei-Fei}]{b13}
Deng, J.; Dong, W.; Socher, R.; Li, L.-J.; Li, K.; and Fei-Fei, L. 2009.
\newblock Imagenet: A large-scale hierarchical image database.
\newblock In \emph{2009 IEEE conference on computer vision and pattern
  recognition}, 248--255. Ieee.

\bibitem[{Dong et~al.(2022)Dong, Bao, Chen, Zhang, Yu, Yuan, Chen, and
  Guo}]{b21}
Dong, X.; Bao, J.; Chen, D.; Zhang, W.; Yu, N.; Yuan, L.; Chen, D.; and Guo, B.
  2022.
\newblock Cswin transformer: A general vision transformer backbone with
  cross-shaped windows.
\newblock In \emph{Proceedings of the IEEE/CVF Conference on Computer Vision
  and Pattern Recognition}, 12124--12134.

\bibitem[{Dosovitskiy et~al.(2020)Dosovitskiy, Beyer, Kolesnikov, Weissenborn,
  Zhai, Unterthiner, Dehghani, Minderer, Heigold, Gelly et~al.}]{b4}
Dosovitskiy, A.; Beyer, L.; Kolesnikov, A.; Weissenborn, D.; Zhai, X.;
  Unterthiner, T.; Dehghani, M.; Minderer, M.; Heigold, G.; Gelly, S.; et~al.
  2020.
\newblock An image is worth 16x16 words: Transformers for image recognition at
  scale.
\newblock \emph{arXiv preprint arXiv:2010.11929}.

\bibitem[{Engelmore and Morgan(1986)}]{em:86}
Engelmore, R.; and Morgan, A., eds. 1986.
\newblock \emph{Blackboard Systems}.
\newblock Reading, Mass.: Addison-Wesley.

\bibitem[{Fan et~al.(2021)Fan, Xiong, Mangalam, Li, Yan, Malik, and
  Feichtenhofer}]{b5}
Fan, H.; Xiong, B.; Mangalam, K.; Li, Y.; Yan, Z.; Malik, J.; and
  Feichtenhofer, C. 2021.
\newblock Multiscale vision transformers.
\newblock In \emph{Proceedings of the IEEE/CVF international conference on
  computer vision}, 6824--6835.

\bibitem[{Guo et~al.(2022)Guo, Xu, Liu, Liu, Jiang, Mu, Zhang, Martin, Cheng,
  and Hu}]{b1}
Guo, M.-H.; Xu, T.-X.; Liu, J.-J.; Liu, Z.-N.; Jiang, P.-T.; Mu, T.-J.; Zhang,
  S.-H.; Martin, R.~R.; Cheng, M.-M.; and Hu, S.-M. 2022.
\newblock Attention mechanisms in computer vision: A survey.
\newblock \emph{Computational visual media}, 8(3): 331--368.

\bibitem[{Han et~al.(2022)Han, Wang, Chen, Chen, Guo, Liu, Tang, Xiao, Xu, Xu
  et~al.}]{b20}
Han, K.; Wang, Y.; Chen, H.; Chen, X.; Guo, J.; Liu, Z.; Tang, Y.; Xiao, A.;
  Xu, C.; Xu, Y.; et~al. 2022.
\newblock A survey on vision transformer.
\newblock \emph{IEEE transactions on pattern analysis and machine
  intelligence}, 45(1): 87--110.

\bibitem[{Hasling, Clancey, and Rennels(1984)}]{hcr:83}
Hasling, D.~W.; Clancey, W.~J.; and Rennels, G. 1984.
\newblock Strategic explanations for a diagnostic consultation system.
\newblock \emph{International Journal of Man-Machine Studies}, 20(1): 3--19.

\bibitem[{Hasling et~al.(1983)Hasling, Clancey, Rennels, and Test}]{hcrt:83}
Hasling, D.~W.; Clancey, W.~J.; Rennels, G.~R.; and Test, T. 1983.
\newblock {Strategic Explanations in Consultation---Duplicate}.
\newblock \emph{The International Journal of Man-Machine Studies}, 20(1):
  3--19.

\bibitem[{He et~al.(2022)He, Chen, Xie, Li, Doll{\'a}r, and Girshick}]{b24}
He, K.; Chen, X.; Xie, S.; Li, Y.; Doll{\'a}r, P.; and Girshick, R. 2022.
\newblock Masked autoencoders are scalable vision learners.
\newblock In \emph{Proceedings of the IEEE/CVF conference on computer vision
  and pattern recognition}, 16000--16009.

\bibitem[{He et~al.(2016)He, Zhang, Ren, and Sun}]{b6}
He, K.; Zhang, X.; Ren, S.; and Sun, J. 2016.
\newblock Deep residual learning for image recognition.
\newblock In \emph{Proceedings of the IEEE conference on computer vision and
  pattern recognition}, 770--778.

\bibitem[{Hendrycks and Gimpel(2016)}]{b27}
Hendrycks, D.; and Gimpel, K. 2016.
\newblock Gaussian error linear units (gelus).
\newblock \emph{arXiv preprint arXiv:1606.08415}.

\bibitem[{Hoffer et~al.(2020)Hoffer, Ben-Nun, Hubara, Giladi, Hoefler, and
  Soudry}]{b37}
Hoffer, E.; Ben-Nun, T.; Hubara, I.; Giladi, N.; Hoefler, T.; and Soudry, D.
  2020.
\newblock Augment your batch: Improving generalization through instance
  repetition.
\newblock In \emph{Proceedings of the IEEE/CVF Conference on Computer Vision
  and Pattern Recognition}, 8129--8138.

\bibitem[{Iandola et~al.(2014)Iandola, Moskewicz, Karayev, Girshick, Darrell,
  and Keutzer}]{b17}
Iandola, F.; Moskewicz, M.; Karayev, S.; Girshick, R.; Darrell, T.; and
  Keutzer, K. 2014.
\newblock Densenet: Implementing efficient convnet descriptor pyramids.
\newblock \emph{arXiv preprint arXiv:1404.1869}.

\bibitem[{Kirillov et~al.(2023)Kirillov, Mintun, Ravi, Mao, Rolland, Gustafson,
  Xiao, Whitehead, Berg, Lo et~al.}]{b2}
Kirillov, A.; Mintun, E.; Ravi, N.; Mao, H.; Rolland, C.; Gustafson, L.; Xiao,
  T.; Whitehead, S.; Berg, A.~C.; Lo, W.-Y.; et~al. 2023.
\newblock Segment anything.
\newblock \emph{arXiv preprint arXiv:2304.02643}.

\bibitem[{Krizhevsky, Hinton et~al.(2009)}]{b40}
Krizhevsky, A.; Hinton, G.; et~al. 2009.
\newblock Learning multiple layers of features from tiny images.

\bibitem[{Krizhevsky, Sutskever, and Hinton(2012)}]{b7}
Krizhevsky, A.; Sutskever, I.; and Hinton, G.~E. 2012.
\newblock Imagenet classification with deep convolutional neural networks.
\newblock \emph{Advances in neural information processing systems}, 25.

\bibitem[{Li et~al.(2022)Li, Mao, Girshick, and He}]{b41}
Li, Y.; Mao, H.; Girshick, R.; and He, K. 2022.
\newblock Exploring plain vision transformer backbones for object detection.
\newblock In \emph{European Conference on Computer Vision}, 280--296. Springer.

\bibitem[{Lin et~al.(2014)Lin, Maire, Belongie, Hays, Perona, Ramanan,
  Doll{\'a}r, and Zitnick}]{b43}
Lin, T.-Y.; Maire, M.; Belongie, S.; Hays, J.; Perona, P.; Ramanan, D.;
  Doll{\'a}r, P.; and Zitnick, C.~L. 2014.
\newblock Microsoft coco: Common objects in context.
\newblock In \emph{Computer Vision--ECCV 2014: 13th European Conference,
  Zurich, Switzerland, September 6-12, 2014, Proceedings, Part V 13}, 740--755.
  Springer.

\bibitem[{Liu et~al.(2021)Liu, Lin, Cao, Hu, Wei, Zhang, Lin, and Guo}]{b8}
Liu, Z.; Lin, Y.; Cao, Y.; Hu, H.; Wei, Y.; Zhang, Z.; Lin, S.; and Guo, B.
  2021.
\newblock Swin transformer: Hierarchical vision transformer using shifted
  windows.
\newblock In \emph{Proceedings of the IEEE/CVF international conference on
  computer vision}, 10012--10022.

\bibitem[{Loshchilov and Hutter(2017)}]{b36}
Loshchilov, I.; and Hutter, F. 2017.
\newblock Decoupled weight decay regularization.
\newblock \emph{arXiv preprint arXiv:1711.05101}.

\bibitem[{Luo et~al.(2016{\natexlab{a}})Luo, Li, Urtasun, and Zemel}]{b29}
Luo, W.; Li, Y.; Urtasun, R.; and Zemel, R. 2016{\natexlab{a}}.
\newblock Understanding the effective receptive field in deep convolutional
  neural networks.
\newblock \emph{Advances in neural information processing systems}, 29.

\bibitem[{Luo et~al.(2016{\natexlab{b}})Luo, Li, Urtasun, and Zemel}]{b42}
Luo, W.; Li, Y.; Urtasun, R.; and Zemel, R. 2016{\natexlab{b}}.
\newblock Understanding the effective receptive field in deep convolutional
  neural networks.
\newblock \emph{Advances in neural information processing systems}, 29.

\bibitem[{{NASA}(2015)}]{c:23}
{NASA}. 2015.
\newblock Pluto: The 'Other' Red Planet.
\newblock \url{https://www.nasa.gov/nh/pluto-the-other-red-planet}.
\newblock Accessed: 2018-12-06.

\bibitem[{Paszke et~al.(2019)Paszke, Gross, Massa, Lerer, Bradbury, Chanan,
  Killeen, Lin, Gimelshein, Antiga et~al.}]{b32}
Paszke, A.; Gross, S.; Massa, F.; Lerer, A.; Bradbury, J.; Chanan, G.; Killeen,
  T.; Lin, Z.; Gimelshein, N.; Antiga, L.; et~al. 2019.
\newblock Pytorch: An imperative style, high-performance deep learning library.
\newblock \emph{Advances in neural information processing systems}, 32.

\bibitem[{Polyak and Juditsky(1992)}]{b38}
Polyak, B.~T.; and Juditsky, A.~B. 1992.
\newblock Acceleration of stochastic approximation by averaging.
\newblock \emph{SIAM journal on control and optimization}, 30(4): 838--855.

\bibitem[{Rice(1986)}]{r:86}
Rice, J. 1986.
\newblock {Poligon: A System for Parallel Problem Solving}.
\newblock Technical Report KSL-86-19, Dept.\ of Computer Science, Stanford
  Univ.

\bibitem[{Robinson(1980{\natexlab{a}})}]{r:80}
Robinson, A.~L. 1980{\natexlab{a}}.
\newblock New Ways to Make Microcircuits Smaller.
\newblock \emph{Science}, 208(4447): 1019--1022.

\bibitem[{Robinson(1980{\natexlab{b}})}]{r:80x}
Robinson, A.~L. 1980{\natexlab{b}}.
\newblock {New Ways to Make Microcircuits Smaller---Duplicate Entry}.
\newblock \emph{Science}, 208: 1019--1026.

\bibitem[{Rogozhnikov(2022)}]{b31}
Rogozhnikov, A. 2022.
\newblock Einops: Clear and Reliable Tensor Manipulations with Einstein-like
  Notation.
\newblock In \emph{International Conference on Learning Representations}.

\bibitem[{Sanders and Kandrot(2010)}]{b33}
Sanders, J.; and Kandrot, E. 2010.
\newblock \emph{CUDA by example: an introduction to general-purpose GPU
  programming}.
\newblock Addison-Wesley Professional.

\bibitem[{Simonyan and Zisserman(2014)}]{b15}
Simonyan, K.; and Zisserman, A. 2014.
\newblock Very deep convolutional networks for large-scale image recognition.
\newblock \emph{arXiv preprint arXiv:1409.1556}.

\bibitem[{Szegedy et~al.(2015)Szegedy, Liu, Jia, Sermanet, Reed, Anguelov,
  Erhan, Vanhoucke, and Rabinovich}]{b16}
Szegedy, C.; Liu, W.; Jia, Y.; Sermanet, P.; Reed, S.; Anguelov, D.; Erhan, D.;
  Vanhoucke, V.; and Rabinovich, A. 2015.
\newblock Going deeper with convolutions.
\newblock In \emph{Proceedings of the IEEE conference on computer vision and
  pattern recognition}, 1--9.

\bibitem[{Tan and Le(2019)}]{b19}
Tan, M.; and Le, Q. 2019.
\newblock Efficientnet: Rethinking model scaling for convolutional neural
  networks.
\newblock In \emph{International conference on machine learning}, 6105--6114.
  PMLR.

\bibitem[{Theodoridis and Koutroumbas(2006)}]{b45}
Theodoridis, S.; and Koutroumbas, K. 2006.
\newblock \emph{Pattern recognition}.
\newblock Elsevier.

\bibitem[{Tolstikhin et~al.(2021)Tolstikhin, Houlsby, Kolesnikov, Beyer, Zhai,
  Unterthiner, Yung, Steiner, Keysers, Uszkoreit et~al.}]{b26}
Tolstikhin, I.~O.; Houlsby, N.; Kolesnikov, A.; Beyer, L.; Zhai, X.;
  Unterthiner, T.; Yung, J.; Steiner, A.; Keysers, D.; Uszkoreit, J.; et~al.
  2021.
\newblock Mlp-mixer: An all-mlp architecture for vision.
\newblock \emph{Advances in neural information processing systems}, 34:
  24261--24272.

\bibitem[{Touvron et~al.(2021)Touvron, Cord, Douze, Massa, Sablayrolles, and
  J{\'e}gou}]{b35}
Touvron, H.; Cord, M.; Douze, M.; Massa, F.; Sablayrolles, A.; and J{\'e}gou,
  H. 2021.
\newblock Training data-efficient image transformers \& distillation through
  attention.
\newblock In \emph{International conference on machine learning}, 10347--10357.
  PMLR.

\bibitem[{Vaswani et~al.(2017{\natexlab{a}})Vaswani, Shazeer, Parmar,
  Uszkoreit, Jones, Gomez, Kaiser, and Polosukhin}]{c:22}
Vaswani, A.; Shazeer, N.; Parmar, N.; Uszkoreit, J.; Jones, L.; Gomez, A.~N.;
  Kaiser, L.; and Polosukhin, I. 2017{\natexlab{a}}.
\newblock Attention Is All You Need.
\newblock arXiv:1706.03762.

\bibitem[{Vaswani et~al.(2017{\natexlab{b}})Vaswani, Shazeer, Parmar,
  Uszkoreit, Jones, Gomez, Kaiser, and Polosukhin}]{b12}
Vaswani, A.; Shazeer, N.; Parmar, N.; Uszkoreit, J.; Jones, L.; Gomez, A.~N.;
  Kaiser, {\L}.; and Polosukhin, I. 2017{\natexlab{b}}.
\newblock Attention is all you need.
\newblock \emph{Advances in neural information processing systems}, 30.

\bibitem[{Wang et~al.(2020)Wang, Sun, Cheng, Jiang, Deng, Zhao, Liu, Mu, Tan,
  Wang et~al.}]{b18}
Wang, J.; Sun, K.; Cheng, T.; Jiang, B.; Deng, C.; Zhao, Y.; Liu, D.; Mu, Y.;
  Tan, M.; Wang, X.; et~al. 2020.
\newblock Deep high-resolution representation learning for visual recognition.
\newblock \emph{IEEE transactions on pattern analysis and machine
  intelligence}, 43(10): 3349--3364.

\bibitem[{Wang et~al.(2021)Wang, Xie, Li, Fan, Song, Liang, Lu, Luo, and
  Shao}]{b22}
Wang, W.; Xie, E.; Li, X.; Fan, D.-P.; Song, K.; Liang, D.; Lu, T.; Luo, P.;
  and Shao, L. 2021.
\newblock Pyramid vision transformer: A versatile backbone for dense prediction
  without convolutions.
\newblock In \emph{Proceedings of the IEEE/CVF international conference on
  computer vision}, 568--578.

\bibitem[{Weems et~al.(1989)Weems, Levitan, Hanson, Riseman, Shu, and
  Nash}]{b46}
Weems, C.~C.; Levitan, S.~P.; Hanson, A.~R.; Riseman, E.~M.; Shu, D.~B.; and
  Nash, J.~G. 1989.
\newblock The image understanding architecture.
\newblock \emph{International Journal of computer vision}, 2: 251--282.

\bibitem[{Werbos(1994)}]{b34}
Werbos, P.~J. 1994.
\newblock \emph{The roots of backpropagation: from ordered derivatives to
  neural networks and political forecasting}, volume~1.
\newblock John Wiley \& Sons.

\bibitem[{Wightman(2019)}]{b39}
Wightman, R. 2019.
\newblock PyTorch Image Models.
\newblock \url{https://github.com/rwightman/pytorch-image-models}.

\bibitem[{Wu et~al.(2021)Wu, Xiao, Codella, Liu, Dai, Yuan, and Zhang}]{b3}
Wu, H.; Xiao, B.; Codella, N.; Liu, M.; Dai, X.; Yuan, L.; and Zhang, L. 2021.
\newblock Cvt: Introducing convolutions to vision transformers.
\newblock In \emph{Proceedings of the IEEE/CVF international conference on
  computer vision}, 22--31.

\bibitem[{Xie et~al.(2022)Xie, Zhang, Cao, Lin, Bao, Yao, Dai, and Hu}]{b25}
Xie, Z.; Zhang, Z.; Cao, Y.; Lin, Y.; Bao, J.; Yao, Z.; Dai, Q.; and Hu, H.
  2022.
\newblock Simmim: A simple framework for masked image modeling.
\newblock In \emph{Proceedings of the IEEE/CVF Conference on Computer Vision
  and Pattern Recognition}, 9653--9663.

\bibitem[{Yu and Koltun(2015)}]{b10}
Yu, F.; and Koltun, V. 2015.
\newblock Multi-scale context aggregation by dilated convolutions.
\newblock \emph{arXiv preprint arXiv:1511.07122}.

\bibitem[{Zhou et~al.(2017)Zhou, Zhao, Puig, Fidler, Barriuso, and
  Torralba}]{b44}
Zhou, B.; Zhao, H.; Puig, X.; Fidler, S.; Barriuso, A.; and Torralba, A. 2017.
\newblock Scene parsing through ade20k dataset.
\newblock In \emph{Proceedings of the IEEE conference on computer vision and
  pattern recognition}, 633--641.

\end{thebibliography}
\clearpage

\appendix
\section{Appendix}
\subsection{MSA-Conv Implementation Details}

In this paper, we propose the MSA-Conv, which involves extracting local regions of $K\times K$ from the input token map. And then applies self-attention to these local regions and extends the computation to the entire map through a sliding window mechanism, effectively combining global and local information~(i.e as the model layers become deeper, TiC can obtain a global receptive field as ResNet \cite{b11}). During implementation, we initially explored using PyTorch's Unfold function \cite{b32} for extracting the sliding window-like values but found it computationally inefficient for large feature maps. To overcome this, we adopted C++ for CUDA programming \cite{b33} to efficiently implement MSA-Conv's forward and backward propagation, significantly reducing computational consumption. Specifically, we implement our MSA-Conv in two parts: (1)~$\mathcal{O}_{QK}$: use $\mathbf{Q}$ and $\mathbf{K}$ to compute $\mathbf{Dot}$, (2) $\mathcal{O}_{DV}$: use \textbf{Dot} and \textbf{V} to calculate the outputs, and the forward propagation is formulated as:

\begin{equation}
    \label{eq:equation_3}
    \begin{aligned}
    & \mathcal{O}_{QK} = \frac{\mathbf{Q} \times \mathbf{K}^{T}}{\sqrt{d^{k}}} \\
    & \mathbf{Dot} = softmax(\mathcal{O}_{QK}) \\
    & \mathcal{O}_{DV} = \mathbf{Dot} \times \mathbf{V}
    \end{aligned}
\end{equation}

\noindent where $d^{k}$ represents the length of the vector, $\mathcal{O}_{QK}$ and $\mathcal{O}_{DV}$ are based on CUDA programming, and $softmax$ are directly called from pytorch interface \cite{b32}. For the specific symbolic derivation of the backward, since we have divided the implementation into two parts, $\mathcal{O}_{QK}$ and $\mathcal{O}_{DV}$, we need to derive the gradient of $\mathbf{Q}$, $\mathbf{K}$, $\mathbf{Dot}$, and $\mathbf{V}$respectively, which is also divided into two parts. The formulation of the backward propagation is defined as:

\begin{equation}
    \label{eq:equation_4}
    \begin{aligned}
    & \frac{\partial \mathcal{L}}{\partial \mathbf{Q}_{b,c,w,h}}=\sum_{i=0}^{K-1}\sum_{j=0}^{K-1} Unfold(\mathbf{K})_{b,i,j,c,w,h} \\
    & \frac{\partial \mathcal{L}}{\partial \mathbf{K}_{b,c,h,w}}=\sum_{i=0}^{K-1}\sum_{j=0}^{K-1} q_{b, c, i_{1}, j_{1}} \cdot \mathbf{\tilde{O}}_{k-1-a,k-1-b,i_{1}, j_{1}} \\
    & \frac{\partial \mathcal{L}}{\partial \mathbf{Dot}_{b,k^{2},h,w}}=\sum_{i=0}^{K-1}\sum_{j=0}^{K-1} v_{b, c, i_{1}, j_{1}} \cdot \mathbf{\tilde{O}}_{k-1-a,k-1-b,i_{1}, j_{1}} \\
    & \frac{\partial \mathcal{L}}{\partial \mathbf{V}_{b,c,w,h}}=\sum_{i=0}^{K-1}\sum_{j=0}^{K-1} Unfold(\mathbf{Dot})_{b,i,j,c,w,h} \\
    \end{aligned}
\end{equation}

\noindent where $b,c, w,h$ represent the shape of $ \mathbf{Q}$, $b,k^{2},w,h$ represent the shape of $ \mathbf{Dot}$, $\mathbf{\tilde{O}}$ represents the gradient of the results obtained from the $\mathcal{O}_{DV}$, $i_{1} = h-\frac{k-1}{2}+i$, $j_{1} = w-\frac{k-1}{2}+j$, $Unfold$ represents the PyTorch function \cite{b32}. Finally, we verify the reliability of our backward derivation by comparing the results obtained by the automated gradient derivation of PyTorch \cite{b32} with those obtained by our own defined gradient derivation. The gradient value of the gradient formulation \cite{b34} is defined by:

\begin{equation}
    \label{eq:equation_5}
    \begin{aligned}
    & \frac{\delta f}{\delta x} = Lim_{\Delta \rightarrow0} \frac{f(x+\Delta) - f(x-\Delta)}{2\Delta} \\
    \end{aligned}
\end{equation}

\noindent where $f$ represents the goal function~(e.g. $\mathcal{O}_{QK}$, $\mathcal{O}_{DV}$), $x$ represents the input, and $\Delta = 1e-7$.

\subsection{Architecture Setup}
In this work, we use TiC-B to complete our experiments, and the specific parameters of our TiC-B configuration are shown in Table. \ref{tab:Table.0}:

\begin{table}[!h]
    \centering
    \footnotesize
    \resizebox{8.5cm}{!}{
    \begin{tabular}{c|ccccccc}
    \toprule
        TiC-B & \makecell{Inter \\ Pooling} & \makecell{Kernel \\ Size} & \makecell{Dilation \\ Ratio} & \makecell{Depthwise \\ Size} & \makecell{Dim \\ Layer} & \makecell{Dim \\ Head} & \makecell{Dim \\ Depth} \\ \hline
        Stage1 & 2 & 3 & 4 & 7 & 128 & 4 & 2 \\
        Stage2 & 2 & 3 & 4 & 7 & 256 & 8 & 2 \\
        Stage3 & 1 & 3 & 4 & 5 & 512 & 16 & 18 \\
        Stage4 & 1 & 3 & 4 & 5 & 1024 & 32 & 2 \\
    \bottomrule
    \end{tabular}
    }
    \caption{The architecture of TiC-B.}
    \label{tab:Table.0}
\end{table}

\noindent Benefiting from the flexibility of MSA-Conv, which can set different kernel sizes, the TiC architecture can be designed with many variants, and this paper is mainly based on TiC-B for subsequent experiments.

\begin{figure}[!t]
\centering
\includegraphics[width=0.5\textwidth]{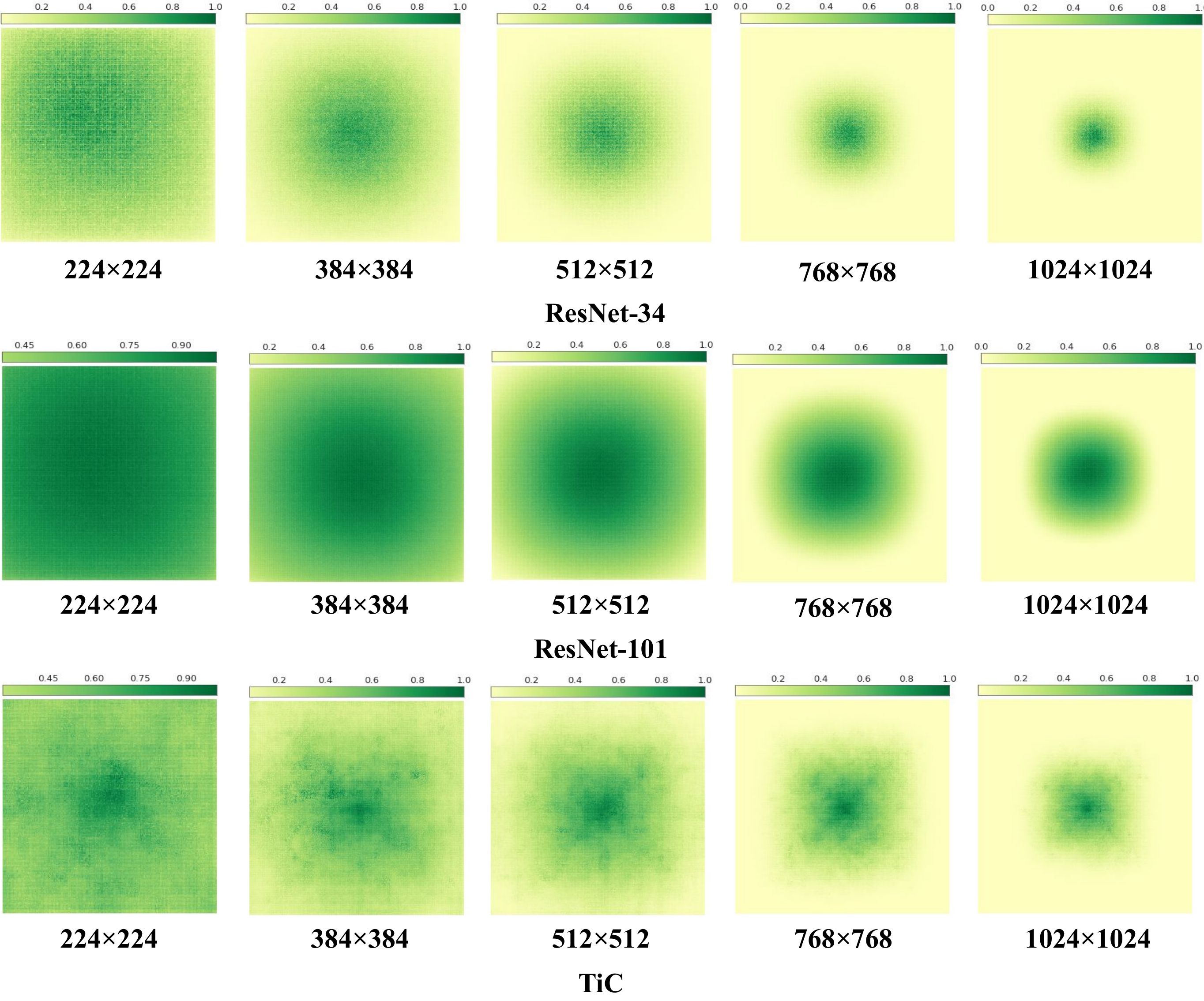}
\caption{Illustration of effective receptive field of ResNet and TiC in various resolution images}
\label{fig:Figure 6}
\end{figure}

\subsection{Modular MSA-Conv Exploratory Experiment}
\label{sec:4.4}
In this section, we further explored the application of modular Multi-Head Self-Attention Convolution in other Conv-Based architectures. As illustrated in Figure \ref{fig:Figure 6}, the utilization of the sliding window mechanism manifests a parallel alteration in the effective receptive field for both our TiC and ResNet architectures as the image scale varies. This congruence in the receptive field behavior signifies a level of interrelation between our TiC model and Convolution-Based architectures. Furthermore, we interfaced MSA-Conv based on pytorch to build a Conv-Like Layer and use it as a convolution layer added to mainstream Conv-Based architectures. Specifically, we added MSA-Conv Layer after the second convolutional layer of Bottleneck in ResNet-18, ResNet-50 \cite{b6}, then tested it on Cifar-100 \cite{b40}, and compared the results with ResNet, TiC and ViT. The experimental results are shown in Table \ref{tab:Table.5}. 

We can observe that for Transformer-Based architectures, due to the amount of data and image information of Cifar-100, it is difficult to train a network with such a large amount of parameters, so it shows low performance on Top-1 acc, e.g. ViT-B: 54.31, TiC-B: 56.88. For Conv-Based architectures, we exploratory add our MSA-Conv to each bottleneck of ResNet18, ResNet50, thus constructing a structure similar to Conv + Attention. It can be seen from the experimental results that thanks to the network structure of Conv-Based architectures, the results of our ResNet18+MSA-Conv, ResNet50+MSA-Conv have a certain performance improvement compared with ViT-B and TiC-B on Cifar-100. In addition, thanks to the mechanism of Conv, our ResNet18+MSA-Conv, ResNet50+MSA-Conv, like ResNet, can accept input of any resolutions without reprocessing the image or configuration. 

In summation, we believe that future research should delve further into modular MSA-Conv and Conv-Based architectures, possibly influencing ViT-based models by synergizing Self-Attention and Convolution for improved outcomes. This has the potential to bolster the model's performance and flexibility.

\begin{table}[!t]
    \centering
    \footnotesize
    \begin{tabular}{c|ccc}
    \toprule
        Methods & \makecell{Image \\ size} & \makecell{Cifar-100 \\ Top-1 acc~(\%)} \\ \hline
        ViT-B & $32^{2}$ & 54.31\\ 
        TiC-B & $32^{2}$ & 56.88\\
        ResNet18 & $32^{2}$ & 75.61  \\
        ResNet18+MSA-Conv & $32^{2}$ & 60.45 \\
        ResNet50 & $32^{2}$ & 77.39\\ 
        ResNet50+MSA-Conv & $32^{2}$ & 61.91\\ 
    \bottomrule
    \end{tabular}
    \caption{Comparison of different backbones pretraining on Cifar-100 \cite{b40}}
    \label{tab:Table.5}
\end{table}

\end{document}